# Where can I drive?
# A System Approach: Deep Ego Corridor Estimation for Robust Automated Driving

Thomas Michalke, Di Feng, Claudius Gläser and Fabian Timm

*Abstract*—Lane detection is an essential part of the perception sub-architecture of any automated driving (AD) or advanced driver assistance system (ADAS). When focusing on low-cost, large scale products for automated driving, model-driven approaches for the detection of lane markings have proven good performance. More recently, data-driven approaches have been proposed that target the drivable area / freespace mainly in inner-city applications. Focus of these approaches is less on lane-based driving due to the fact that the lane concept does not fully apply in unstructured, residential inner-city environments. So-far the concept of drivable area is seldom used for highway and inter-urban applications due to the specific requirements of these scenarios that require clear lane associations of all traffic participants. We believe that lane-based, mapless driving in inter-urban and highway scenarios is still not fully handled with sufficient robustness and availability. Especially for challenging weather situations such as heavy rain, fog, low-standing sun, darkness or reflections in puddles, the mapless detection of lane markings decreases significantly or completely fails. We see potential in applying specifically designed data-driven freespace approaches in more lane-based driving applications for highways and inter-urban use. Therefore, we propose to classify specifically a drivable corridor of the ego lane on pixel level with a deep learning approach. Our approach is kept computationally efficient with only 0.66 million parameters allowing its application in large scale products. Thus, we were able to easily integrate into an online AD system of a test vehicle. We demonstrate the performance of our approach under challenging conditions qualitatively and quantitatively in comparison to a state-of-the-art model-driven approach. We see the current approach as part of a fallback path whenever model-driven approaches cannot cope with a challenging scenario. We give insights how such a fallback path can be integrated into an AD system, thereby extending the overall system availability.

## I. Introduction

Traffic accidents are still a major problem and currently the eighth leading fatality in the world. According to a recent report of the World Health Organization, traffic accidents continue to claim more than 1.3 million lives and cause about 50 million injuries every year [1]. Many of these accidents occur because the driver is inattentive and the vehicle leaves the lane [2]. Developments in the field of driver assistance (DA), such as the Lane Keeping Assist System and the Lane Departure Warning System, show a high potential to decrease the number of accidents [3], [4], [5]. Therefore, reliable and robust lane detection is essential and must be further improved, especially for future AD systems [6][4] that will require a high coverage and system availability.

There are many different solutions to lane detection, which we divide into three categories: Data from stationary Video and/or Lidar sensors, data from high-definition maps and data from the vehicle's Video and/or Lidar sensors. Although data of infrastructure sensors has some advantages such as easier calibration or no ego-motion compensation, its street coverage is still very limited. Therefore, this data may only be used at accident black spots and inner cities. Data from high-definition maps can overcome many drawbacks of other sensor modalities. However, this data is not available everywhere and must be verified and updated regularly. Due to safety requirements w.r.t redundancy as well as performance and availability, a mapless perception for driving lane detection is essential.

Reliable and robust lane detection with onboard sensors must overcome several challenges [7][8]:

- Partially faded or missing road markings on new, but yet unfinished roads,
- Shadows by other vehicles, buildings or vegetation,
- Text and other symbols on the road, such as speed limits,
- Severe environmental conditions like heavy rain or reflections from low sunlight.

We propose an efficient deep learning approach for ego-lane detection, see Fig. 2, which yields accurate and reliable results even in very challenging situations. We show that it can be robustly integrated into an AD system, see Fig. 1 and applied in real-time in a prototype vehicle.

## II. Related Work

Lane detection has been studied for more than two decades [2] and a large variety of approaches has been proposed [20]. Traditional methods treat lane detection as a line fitting problem [21]. Thereby, they apply a two-stage regime in which edges are detected first, before geometrical models are subsequently fitted via Hough transform, dynamic programming, or energy minimization approaches [22], [23], [24], [25]. As a predecessor of recent learning-based approaches, feature-driven approaches with manually selected image features have shown better performance in unstructured environments and under challenging weather conditions [26].

Recently, lane detection approaches in the field of deep learning have shown even more promising results than classical methods. Table I shows some selected state-of-the-art

All authors are with Corporate Research, Robert Bosch GmbH, 71272 Renningen, Germany `Firstname.Lastname@de.bosch.com`.

TABLE I: Selected state-of-the-art approaches for lane detection - we only consider references of already published articles, i.e. we ignore results of anonymous submissions to public benchmark data sets. Moreover, if the architecture was not fully defined we denote the number of parameters by "—".

| Method | Year | Ref. | Approach | Input | #Params [Mio] |
|---|---|---|---|---|---|
| ELCNN | 2017 | [9] | CNN + RPN | 752×480 | — |
| SCRFFPFHGSP | 2015 | [10] | compos. high order pattern potentials | 640×480×2 (stereo) | — |
| SPRAY | 2012 | [11] | ray features + boost classifier | 800×600×3 | — |
| RBNet | 2017 | [12] | Bayesian CNN | 900×300 | — |
| Multi-camera | 2016 | [13] | RNN + LSTM | 500×80 stripes | — |
| VPGNet | 2017 | [14] | CNN | 640×480×3 | 301.00 |
| Up-Conv-Poly | 2016 | [15] | CNN | 500×500 | 19.44 |
| StixelNet | 2015 | [16] | CNN + CRF | 24×370×3 stripes | 6.82 |
| Multi-sensor | 2018 | [17] | CNN | 960×960 BEV | 5.70 |
| LaneNet | 2018 | [18] | CNN | 512×256 | 0.56 |
| RoadNet3 | 2019 | [19] | CNN+LSTM | 600×160×5 | 0.36 |
| **Our approach** | | | CNN | 652×360 | 0.66 |

methods for lane detection. Many of them treat lane detection as a semantic segmentation task [27]. These algorithms usually apply an encoder-decoder network structure that first reduces the spatial resolution, extracts high-level features and subsequently upsamples the representation to obtain pixel-wise labels in the target image or birds-eye view resolution [15], [17]. Recent methods further employ multi-task models in which lane detection is embedded into a network covering other perception tasks like obstacle detection, road segmentation, or road marking detection [14], [16], [18]. Most aforementioned approaches only use a single image as input, whereas others propose hybrid networks that combine the encoder-decoder structure with recurrent LSTM layers to leverage temporal information for robust lane detection and road segmentation [19], [28].

Furthermore, there are various approaches which first fuse sequentially arising sensor data to obtain temporally filtered representations of the vehicle surrounding. These methods most often employ occupancy grid maps for sensor data fusion based on which the road course, the road boundary, or the ego-vehicle corridor is subsequently predicted [29], [30], [31], [32].

From a safety-driven perspective, classical model-driven approaches are still a valid option even for AD systems. These approaches show a good performance in structured environments under normal weather conditions. Therefore in such situations the approach is sufficient for the requirements that modern AD systems pose. Furthermore, their situative shortcomings are well understood. For these situations, independent perception paths can take over as a fallback solution. Thereby a one fits all approach is not required facilitating the task of finding a safe, performant, and highly available system solution.

The main contributions of this paper are:
- Network specifically trained to handle challenging situations for lane-based driving (inter-urban, highways),
- Redundant fallback path addressing failure cases of classical line perception,
- Small network running in real-time in a test vehicle with full loop closure (i.e. with active control interventions).

## III. SYSTEM

In contrast to the previously mentioned approaches, the goal of this paper is NOT to perform a classical semantic segmentation for detecting the freespace in an input image (e.g. the oncoming lane is also freespace). Instead, we aim to add more semantics to the semantic segmentation approach by predicting the drivable corridor of the ego-lane including dynamic scene content as the preceding vehicle. The detected corridor is hence restricted by the preceding vehicle to the front as well as the corridor borders to the side (see Figure 5). Both information can be used for active vehicle control (longitudinal and lateral). In this paper, we focus on its application for lateral control.

### A. System Overview

Figure 1 provides a rough overview of a state-of-the-art AD system with a more thorough focus on the perception modules required for lane-based driving (e.g. in structured inter-urban and highway scenarios). Typically, AD stacks are clustered into the following sub-architectures: *Sensors*, *map*, *perception and environment modeling*, *situation analysis*, *behavior and motion planning*, *motion control* and *actuators*. As *sensors* for automated vehicles multiple redundant sensor modalities are applied: Radar, monocular and stereo video sensors, lidars as well as car communication unit allowing for connectivity to other vehicles or infrastructure (V2X communication). Highly-precise *maps* are applied in combination with GPS- and sensor-driven road signatures for precise global localization. Within the sub-architecture for *perception and environment modelling* the static and dynamic environment is detected, tracked, fused and represented. Typically, object lists are used for structured scene content (e.g. vehicles, pedestrians, line markings). Additionally freespace and occupancy grids are used as model-free representation in unstructured environments. The *situation analysis* enhances the environment model by reasoning approaches that are driven from context information.

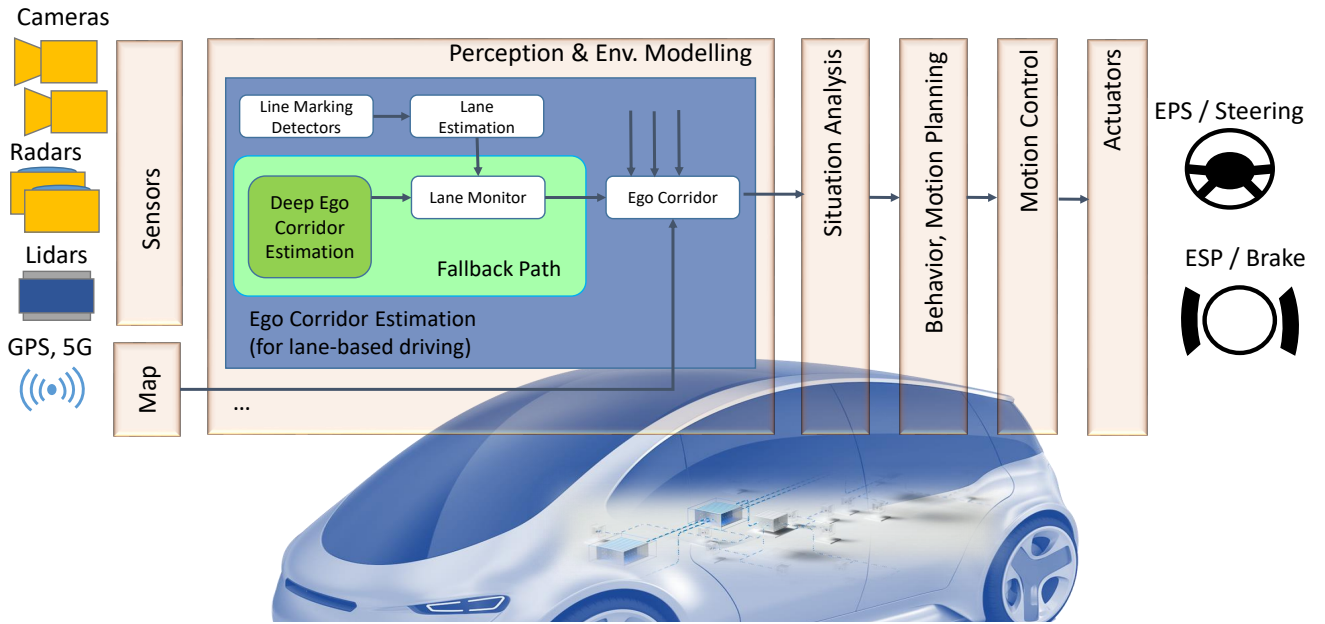

Fig. 1: Overview of state-of-the-art AD system with fallback path for lane marking detection (focus on perception sub-architecture for lane-based driving, weak coupling to HAD map due to safety aspects).

The *planning* sub-architecture determines the high level behavior of the system and based on that computes the desired collision-free trajectory typically relying on plattform-independent motion planning algorithms. The *motion control* sub-architecture provides the vehicle-specific controllers for lateral and longitudinal control used to steer the vehicle *actuators* (in our case the control units for the Eletronic Power Steering - EPS and the Electronic Stability Programm - ESP). In the following a more detailed description of the perception modules for lane-based driving are provided with a specific focus on the proposed fallback path.

### B. Perception for Ego Corridor Estimating & Fallback Path

Lane-based driving on highways comes with challenging requirements regarding detection range (due to the faster driving velocities) and in case of higher automation levels 3 and 4 also prediction time to allow triggering an early driver take over request in case of situations that violate the system specification. When focusing on lane-based driving on highways two major information sources exist: Highly precise maps and lane markings from video sensors. Radar sensors can provide additional, weak detections for the lane borders (e.g. regression on stationary radar reflections on crash barriers). Lidar sensor can detect lane markings based on relative changes of the reflectivity of the ground, but are restricted in the detection range. Grid approaches can fuse unstructured freespace information, but are also restricted in the decetction range. Independent from the automation level an environment sensor-driven detection approach is first choice due to safety aspects. This means a strong and early combination with map data is no option. Hence as shown in Figure 1 a late fusion between map and video-based lane detections is proposed. For lane detection the most prominent detection methods is a model-based detection of line marking e.g. relying on a Kalman-filter based tracking of the parameters of a clothoid model for the detected lines. In harsh weather conditions these approaches will fail. To assure a high system availability, we hence propose a *fallback path* (see light green surface in Figure 1). Its main component consists of a *lane monitor* module that evaluates the quality of the line marking detections and acts as an intelligent switch between model-based detection and the data-driven approach (see dark green surface in Figure 1). It is designed as a classifier that takes the dynamics and tracking ages of the detected lines into account. Additionally, in order to react in a more predictive manner the *lane monitor* compares statistics of a complex feature vector (e.g. including the state space of behavior-relevant objects, ego dynamics) of the current scenario with the statistics of recorded data of scenarios that matched the spec of the model-driven line detection approach. We use Kernel Density Estimation on the named complex feature vector for efficiently representing the "within spec" data (see [33] for more details on this principle). In case the *lane monitor* classifies the detected lane markings as unstable and the analysis of the current feature vector shows a "outside spec" application, a combination and finally complete switch to the data-driven lane detection approach will be realized. The data-driven approach works as a fall-back path to improve the system availability. For level 3 and 4 automation, the system gains time for a safe driver take-over-request. In a level 2 system, the driver can be supported even in harsh weather conditions, which might be situations especially unexperienced driver will highly benefit from assisted driving.

In the next Section, the proposed data-driven deep ego corridor estimation is introduced in more detail.

## IV. EGO CORRIDOR ESTIMATION

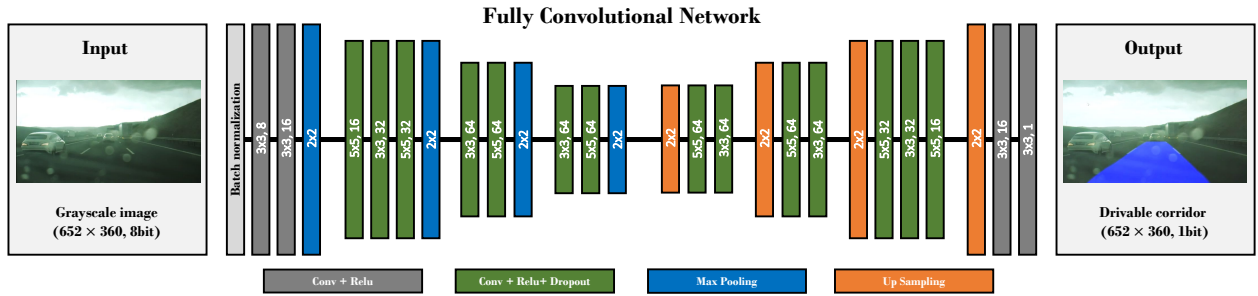

Fig. 2: An overview of our architecture for the detection of a drivable corridor in the ego-lane. For convenience, the output is visualized as overlay in the RGB image.

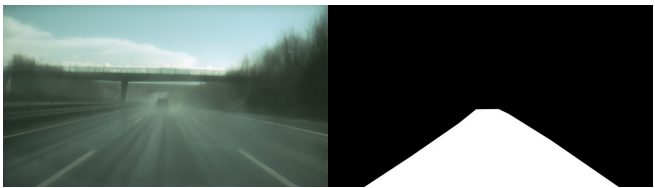

Fig. 3: Labels on the road captured by a real car.

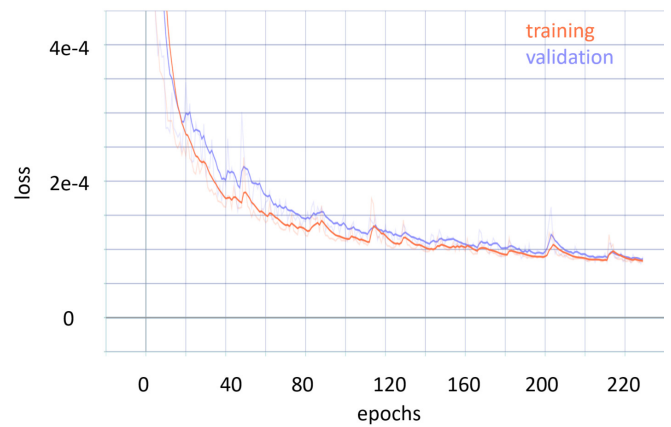

Fig. 4: Loss during training of the neural network (raw values shown as shaded curves, smoothed values as bold curves).

In the following the properties of the module for deep corridor estimation is introduced in more depth.

### A. Creating a Dataset

The ground truth contains roughly 5000 labeled frames in total, which is enough for a proof-of-concept. Additional data will further improve the approach.

The sequences used in the dataset were recorded whenever the lateral control of the test vehicle, using a classical approach for lane detection, was unavailable due to harsh environmental conditions.

Figure 3 shows the labels on the road data with the camera image on the left and the corresponding labeled image on the right.

### B. Neural Network Architecture

The architecture of the neural network is an important factor for both the accuracy of the model as well as the speed of inference. We used the following principles for our model:

- Use a very simple pipeline with only a single network to keep up inference speed and facilitate optimization.
- Keep the network size relatively small, ideally below one million trainable parameters.
- Avoid computationally expensive preprocessing by using the camera image as input data.
- The output data should have the same shape as the input data for easy integration into other systems.
- Compatibility with Tensorflow for integration into the test vehicle.

Figure 2 shows the network architecture that was chosen for this approach. The neural network is implemented in Keras and has a total of 41 layers with a total of approximately 0.66 million trainable parameters. It is a fully convolutional network with a symmetric architecture, meaning that there is a deconvolutional layer for every convolutional layer and an upsampling layer for every pooling layer. This ensures that the output shape remains the same as the input shape. The input is a grayscale image with a resolution of $652 \times 360$ pixels. The network uses a combination of different pooling sizes to increase prediction accuracy while at the same time reducing the number of trainable parameters. A stride of one is used for all convolutional and pooling layers. Rectified Linear Units (ReLU) are used as activation function. The output has the same size as the input providing pseudo probability values between 0 and 1 that encode if the respective pixel belongs to the ego-lane segment.

### C. Neural Network Training

With the dataset and network architecture created, the next step is training the neural network. For this we use Adam [34] as the optimizer and train the network in batches of sixteen frames for a total of 250 epochs. Root mean square error (RMSE) is chosen as the loss function for the optimizer.

Figure 4 shows the loss during training. For the here shown training run, the model reaches an optimum at epoch 223. At the optimum validation and training loss values are very close, which means that there is no clear tendency towards overfitting or underfitting.

## D. Test Vehicle Overview

The test vehicle essentially contains an adaptive cruises control (ACC) system. It is based on radar and uses a mono camera for lateral control. The mono camera is installed in a fixed position aiming at the road ahead. It is able to provide data at a frame rate of 15 Hz. The car uses one Nvidia GTX Titan X graphic card available for processing.

## E. Ensuring Compatibility with the Test Vehicle

All of the chosen technologies must be compatible with the systems in the test vehicle. Thus the available technology choices are somewhat limited.

Additionally, the whole processing pipeline must be real-time capable. In practice, this means that very little preprocessing can be done and that the necessary preprocessing has to be very fast. The time needed for inference of the neural network is not a significant concern due to the small model size and processing power of the systems in the car.

In order to cope with activated windshield wiper interfering with the optical path, a straight-forward method was adopted by computing the mean of the ego-lane detections results of the last three video images.

## V. EXPERIMENTS & RESULTS

In the following, the proposed approach is evaluated in a qualitative and quantitative fashion as well as online in the test vehicle. Additionally, we provide a video of the online performance allowing a better assessment of the proposed approach in dynamic scenes.

## A. Qualitative Results

Figure 5a-h shows the lane detection result overlayed on top of the camera image in a scenarios with sunlight causing reflections on a wet road as well as bad lighting conditions (here exemplarily in a tunnel). The results appear to be accurate, although the lane markings in this scenario are very hard to see for a human as well. Furthermore, Figure 5i-m) shows camera images taken during heavy rain, which is very challenging for lane detection. Lane markings are barely visible due to the water on the road, cars in front cause dense water sprays. Furthermore, raindrops on the windscreen change the optical pathway and distort the view.

## B. Quantitative Results

We are able to measure an average intersection over union (IoU) across a test dataset of 97.6%. The test data set includes roughly 500 labeled frames which were not used for training the neural network. The neural network achieves an inference performance of roughly 70 frames per second (14 ms) in our test vehicle. Performance cost due to preparation of input data and postprocessing for the controller interface are not included in this performance measurement.

In the next step, the performance of the proposed approach and a classical line detection approach based on a Hough transform ([21] [23]) are compared for 17 sequences of challenging situations. The data of these situations was recorded during online closed-loop tests that allowed assessing the restrictions of classical non-AI-based algorithms. Table II shows an overview of the challenge categories present in the dataset, the number of sequences in each category as well as a reference to an example image to allow a better understanding of the challenging situation. Each sequence contains 450 image frames and 1500 frames of vehicle CAN data, which equals to 30 seconds of real-world data. We propose two key performance indicators (KPIs) that relate to the availability of the corridor in front of the vehicle. Based on that, we decide if an approach has the potential to react as required in terms of the Lane Keeping Assist function that depends on the corridor data.

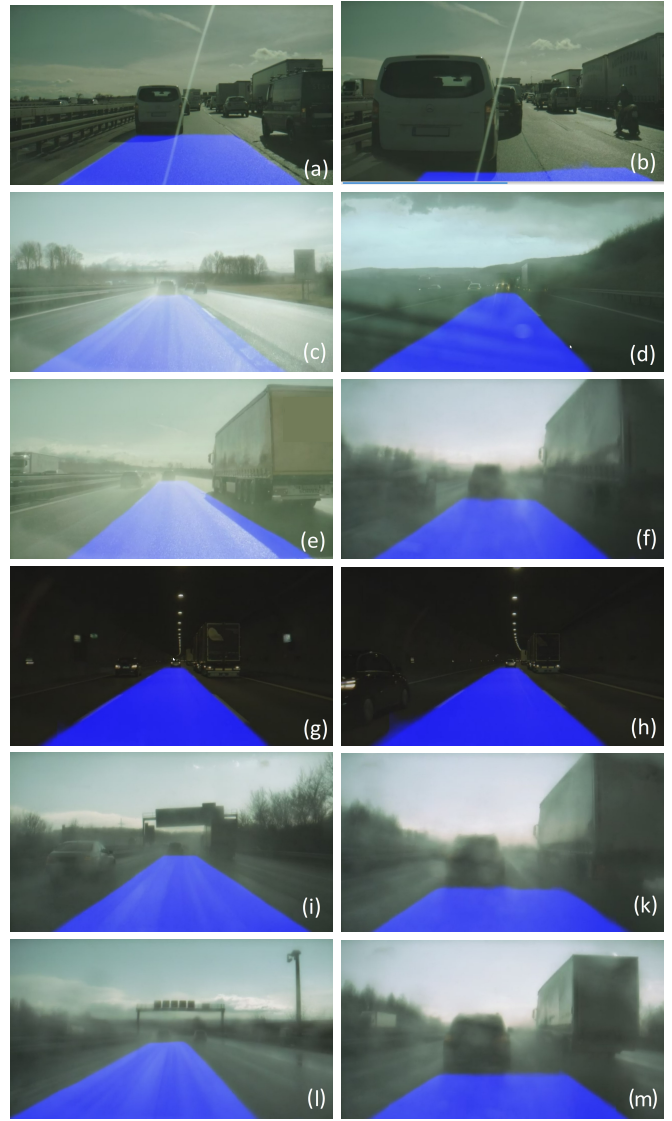

Fig. 5: Results for example images of test data set in challenging situations (direct sunlight, tunnel, heavy rain).

A frame-based KPI is defined as follows:

- Ratio between image frames allowing the activation of automatic lateral control to overall frames in the sequence.

- Automatic lateral control can be activated, when the corridor width is within a band of 2m and 6m and when corridor length is above the 0.7s time-gap distance (simplified assumptions for data sanity check).

A sequence-based KPI is defined as follows:
- Ratio between sequences allowing the activation of automatic lateral control to overall sequences in the category.
- Lateral control is counted as activated for a sequence, if the lateral control is not unavailable for more than 5 consecutive image frames throughout the whole sequence (simplified assumption taking the low-pass characteristic of the steering system into account).

As shown in Table II the proposed approach clearly outperforms the chosen state-of-the-art classical line detection algorithm for challenging situations containing sun after rain, heavy rain, faded lines, and coke cleavages. In situations with direct sunlight, the classical approach performs better. Since data containing direct or indirect sunlight have not been part of the training set, the network generalized still rather well. With an extension of the data set, a significantly improved performance can be expected. The results show that a system that relys on the proposed fallback path will reach an enhanced overall system availability in analyzed challenging situations.

*C. Online Performance in the Test Vehicle*

Finally, we look at the performance of the neural network once integrated into the test vehicle (see Fig. 6 and 7). The provided video and Figure 8 shows the prediction result in the test vehicle on a stretch of highway that has not been part of the test or training dataset. It should be noted that the lighting conditions do not match those present during the training of the neural network. Additionally it shows hard drop shadows from other vehicles and road signs. However, the result still closely matches the performance we have seen during offline evaluation. As shown in Fig. 7, the achieved performance of the here proposed approach allows for hands-free driving. Summarizing, it can be stated that a proof-of-concept was successfully shown based on a rather restricted dataset of 5000 images. Based on that, we will improve and fine-tune to cover all challenging situations that well-approved classical methods cannot cover. Due to its application as fallback solution, the network will keep a restricted number of parameters and hence will stay efficient in terms of memory usage and computation time.

## VI. Conclusion & Outlook

In this paper, an approach for detecting the ego-lane of a vehicle with a fully convolutional network is presented. The proposed network takes the current frame as input and predicts the drivable corridor in the ego-lane as output image.

We successfully applied the approach to real world data captured with a test vehicle including heavy rain and direct sunlight. The neural network is able to significantly improve the lane detection in terms of performance and availability. The ego-lane is detected with an average IoU of 97.6% while maintaining an inference performance of 70 Hz. Finally, the neural network is integrated into a test vehicle for online evaluation on the road.

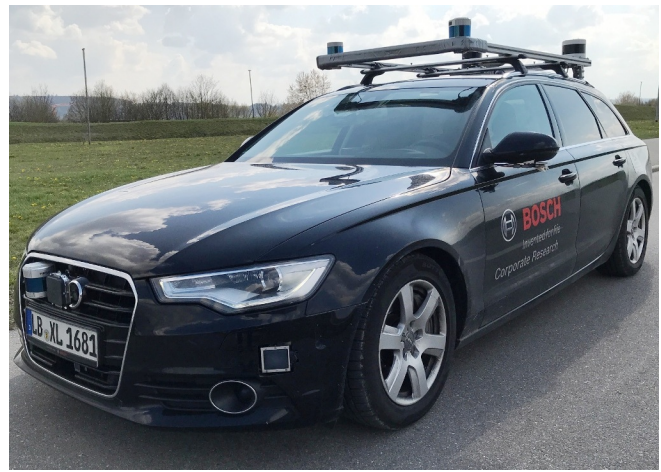

Fig. 6: Test vehicle used for online evaluation.

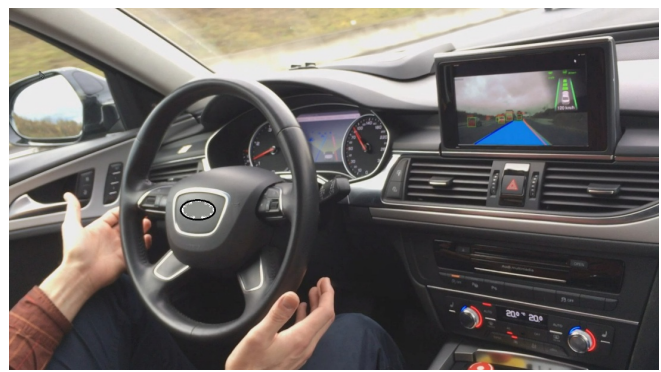

Fig. 7: Online application in the test vehicle (hands have to stay close to the steering wheel due to legal reasons).

The approach is able to consistently deliver accurate lane detection in difficult conditions such as reflections, faded lane markings, heavy rain, poor lighting in tunnels. It is used in a closed-loop system as a fallback for lateral vehicle control.

In the future, we plan to use the length of the detected output lane for longitudinal control of the vehicle. Furthermore, the length of the detected output lane appears to update very quickly when the preceding vehicle cuts into the ego vehicle's lane. We plan to investigate whether this property of the neural network can be used to detect cut-in scenarios earlier than with classical methods.

TABLE II: Availability of a classical lane marking detection approach as compared to the here presented approach.

| Challenges | #Seq. | Classic lane detection | | Our approach | | Example image |
|---|---|---|---|---|---|---|
| | | Frame-based | Seq.-based | Frame-based | Seq.-based | |
| Sun after rain | 2 | 0.21(191/900) | 0/2 | **0.95 (858/900)** | 2/2 | Fig. 5(c-f) |
| Heavy rain | 3 | 0.21(284/1350) | 0/3 | **0.81 (1099/1350)** | 2/3 | Fig. 5(i-m) |
| Direct sunlight | 3 | **0.82 (1109/1350)** | 0/3 | 0.72(967/1350) | 0/3 | Fig. 8(a,c) |
| Faded lines | 6 | 0.44(1182/2700) | 0/6 | **0.75 (2020/2700)** | 4/6 | Fig. 8(b) |
| Coke cleavages | 3 | 0.97(1312/1350) | 2/3 | **0.99 (1339/1350)** | 3/3 | - |

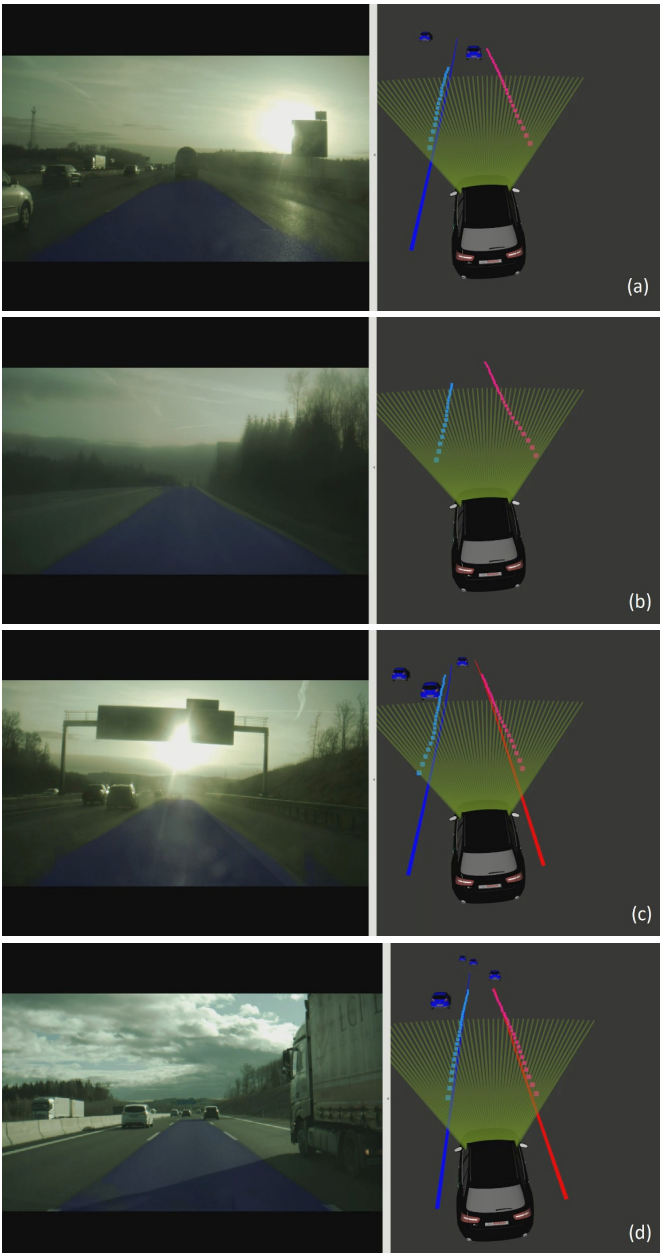

Fig. 8: Perception and controller data recorded during online tests. Left: Detected ML-based ego-lane. Right: Blue and red lines depict the classical marking detection as main perception path. Light blue and pink dots depict the controller input post-processed from the ML-based ego-lane as fallback path. (a,b) Classical lane marking detection cannot cope with the situation, ML-based ego-lane detection remains usable, (c,d) Classical and ML-based ego-lane detection deliver similar usable results.